\pdfoutput=1

\documentclass[11pt]{article}

\usepackage[preprint]{acl}

\usepackage{times}
\usepackage{latexsym}
\usepackage{amsmath}
\usepackage{booktabs}
\usepackage{multirow}
\usepackage{makecell}
\usepackage{tabularx}
\usepackage{algorithm}
\usepackage{algpseudocode}

\usepackage[T1]{fontenc}

\usepackage[utf8]{inputenc}

\usepackage{microtype}

\usepackage{inconsolata}

\usepackage{graphicx}

\title{LIMIT: \textbf{L}ess \textbf{I}s \textbf{M}ore for \textbf{I}nstruction Tuning in \textbf{T}ext-to-SQL}

 \author{
 \textbf{Haoyuan Ma}\textsuperscript{1}, 
 \textbf{Hengwei Liu}\textsuperscript{1},
 \textbf{Linjuan Wu}\textsuperscript{1},
 \textbf{Yongliang Shen}\textsuperscript{1}\textsuperscript{\textdagger}, 
 \textbf{Weiming Lu}\textsuperscript{1}\textsuperscript{\textdagger}  
 \\
$^{1}$Zhejiang University, \\
\texttt{\{mahaoyuan, syl, luwm\}@zju.edu.cn} \\
}

\begin{document}
\maketitle

\renewcommand{\thefootnote}{\textdagger}
\footnotetext{Corresponding author.}

\begin{abstract}

Large language models have achieved remarkable progress on Text-to-SQL through reasoning-enhanced fine-tuning, yet existing approaches predominantly rely on massive instruction corpora under the assumption that scale drives performance.
We challenge this paradigm by investigating a fundamental question: what is the minimal data requirement for effective Text-to-SQL instruction tuning?
We propose \textbf{LIMIT} (\underline{\textbf{L}}ess \underline{\textbf{I}}s \underline{\textbf{M}}ore for \underline{\textbf{I}}nstruction \underline{\textbf{T}}uning in Text-to-SQL), a data-centric framework that demonstrates strong database reasoning can emerge from an extremely compact training set when examples are strategically selected. 
LIMIT operates through four stages: difficulty-aware filtering that identifies samples within the model's learning frontier, chain-of-thought synthesis with consistency-based selection, multi-dimensional quality scoring via LLM-as-judge, and genetic algorithm optimization that jointly maximizes schema coverage and sample quality.
On the BIRD and Spider benchmark, LIMIT selects only 796 and 863 samples while achieving 100\% table coverage, enabling Qwen3-8B to reach 69.1\% and 88.9\% execution accuracy.
This result surpasses methods trained on 20 times more data and establishes a new state-of-the-art among open-source approaches. Our findings suggest that careful data curation, rather than scale, is the key to efficient Text-to-SQL learning.
\end{abstract}

\section{Introduction}


Text-to-SQL systems translate natural language queries into executable database commands, enabling non-experts to interact with structured data without SQL proficiency. Recent advances in large language models \citep{deepseekv3,dapo} have substantially improved performance on this task, with reasoning-enhanced approaches achieving notable success on challenging benchmarks \citep{sql-r1, reasoning-sql}. These methods typically combine SFT (supervised fine-tuning) with RL (reinforcement learning) objectives, leveraging CoT (chain-of-thought) annotations to elicit multi-step reasoning behavior.

A common thread underlying these successes is the reliance on large-scale training data: OmniSQL \citep{omnisql} employs 2.5 million instruction samples, CodeS \citep{codes} requires 21.5GB of pretraining corpora, and even recent reasoning-enhanced approaches \citep{sql-r1, reasoning-sql} depend on tens of thousands of annotated examples. This scaling-centric paradigm, while effective, incurs substantial computational costs and raises questions about whether such data volumes are truly necessary.

Recent work on mathematical reasoning has challenged the conventional wisdom that complex reasoning requires massive datasets. LIMO \citep{limo} demonstrated that when base models have internalized sufficient domain knowledge during pretraining, a small number of high-quality examples can effectively activate latent reasoning capabilities. This finding suggests an intriguing possibility: \textit{can we achieve strong Text-to-SQL performance with dramatically fewer training samples?} 
This is not about filtering out low-quality or irrelevant data, but about identifying the smallest subset of highly valuable examples that offers the greatest instructional benefit.
Answering this question affirmatively would not only reduce computational costs but also provide insights into the fundamental data requirements for structured reasoning tasks.

However, directly transferring the Less-Is-More principle to Text-to-SQL reveals fundamental challenges absent in general reasoning domains. Text-to-SQL demands that models comprehend diverse table structures, column semantics, and referential relationships across heterogeneous databases. A training set achieving high quality scores but covering only partial schema elements will inevitably fail on queries involving unseen tables or columns. This observation exposes a critical tension: minimizing data volume inherently conflicts with maximizing schema coverage, a constraint unique to database-grounded tasks.

We argue that effective data selection for Text-to-SQL must jointly optimize two objectives that previous work treats independently. The first objective concerns database structural coverage, ensuring selected examples collectively span all tables and sufficient columns in target schemas. The second objective concerns instructional quality, requiring each example to exhibit clear semantics, correct SQL implementations, and coherent reasoning traces. Existing selection methods based on difficulty filtering \citep{limo}, diversity sampling \citep{s1}, or quality scoring \citep{omnisql} address only one dimension and therefore achieve suboptimal data efficiency for structured generation tasks.

To address these challenges, we propose \textbf{LIMIT} (\underline{\textbf{L}}ess \underline{\textbf{I}}s \underline{\textbf{M}}ore for \underline{\textbf{I}}nstruction \underline{\textbf{T}}uning in Text-to-SQL), a data-centric framework that systematically constructs minimal yet effective training sets. Our approach operates through four complementary stages. The first stage performs difficulty-aware pre-filtering using a dual-model strategy: samples easily solved by a weak model are removed as uninformative, while samples that a strong model cannot reliably solve are excluded as potentially noisy. This filtering identifies examples within the optimal learning zone. The second stage synthesizes CoT annotations using a reasoning-capable model, selecting the shortest correct reasoning trace to maximize instructional clarity while minimizing redundancy. The third stage employs a unified LLM evaluator to score each candidate along multiple dimensions including question-SQL alignment, input clarity, SQL quality, and reasoning coherence. The fourth stage formulates dataset selection as a combinatorial optimization problem, using a genetic algorithm to search for subsets that jointly maximize schema coverage and aggregate quality while satisfying data budget constraints.

We evaluate LIMIT on the BIRD benchmark \citep{bird}, which features complex real-world databases across diverse domains. Our framework selects only 796 training samples while achieving 100\% table coverage and 86.8\% column coverage. Fine-tuning Qwen3-8B with supervised learning followed by GRPO (group relative policy optimization) \citep{deepseekmath} enables LIMIT to reach 69.1\% execution accuracy, establishing state-of-the-art performance among open-source model approaches. This result is achieved using fewer than 5\% of the training data employed by prior methods, demonstrating that dramatic data reduction is possible without sacrificing performance.

Our contributions are threefold. First, we identify the unique challenges of applying data-efficient learning to Text-to-SQL and formalize the joint optimization of schema coverage and example quality as the central problem. Second, we develop a principled four-stage framework that addresses this optimization through complementary filtering, scoring, and selection mechanisms. Third, we provide empirical evidence that Text-to-SQL models can achieve strong performance with two orders of magnitude less data than previously assumed, suggesting new directions for resource-efficient semantic parsing research.

\section{Related Work}

Early Text-to-SQL efforts built on closed-source LLMs often augmented a frozen backbone with modular components to compensate for limited task-specific adaptation.
Typical modules include schema linking and retrieval to surface relevant tables and columns \citep{RESDSQL, RSL-sql}, decomposition and planning to split complex queries \citep{sgu-sql, metasql, tasql, din-sql}, candidate generation and execution-based ranking \citep{c3, reforce, PET-SQL, mcs-sql, super-sql, sql-plam, AgentarScaleSQLAT}, and iterative execution feedback or multi-agent consensus for self-correction \citep{mac-sql, chess, EPI-SQL}.
While these modules improve robustness without changing model weights, reliance on closed models raises privacy concerns and limits adaptation to database-specific schemas.

Open-source fine-tuning approaches address adaptation and privacy by internalizing similar modules through SFT \citep{dts-sql, db-explore}, continual pretraining on SQL corpora \citep{codes, omnisql}, bidirectional data synthesis \citep{sense} and multi-task data synthesis \citep{ROUTE, slm-sql, XiYan-SQL}, and by adding search or RL objectives (e.g., MCTS, GRPO) to reward executable correctness \citep{sql-o1, sql-r1, reasoning-sql}.
However, like methods in mathematical and reasoning, these pipelines frequently rely on very large collections of annotated, stepwise solutions to stabilize multi-step logical behavior, incurring high annotation and synthesis costs.

The Less-is-More principle \citep{limo} contests that scale is not the only path: a small, carefully curated set of high-difficulty examples with explicit reasoning chains can activate latent reasoning in pretrained models.
Validated across math \citep{s1}, reasoning \citep{lima} and agentic coding tasks \citep{limi}, this data-efficient perspective motivates an alternative tuning approach for the Text-to-SQL task.
Motivated by this, we propose LIMIT, a Less-Is-More instruction-tuning framework that curates training examples to elicit database reasoning.

\begin{figure*}[ht]
  \includegraphics[width=2\columnwidth]{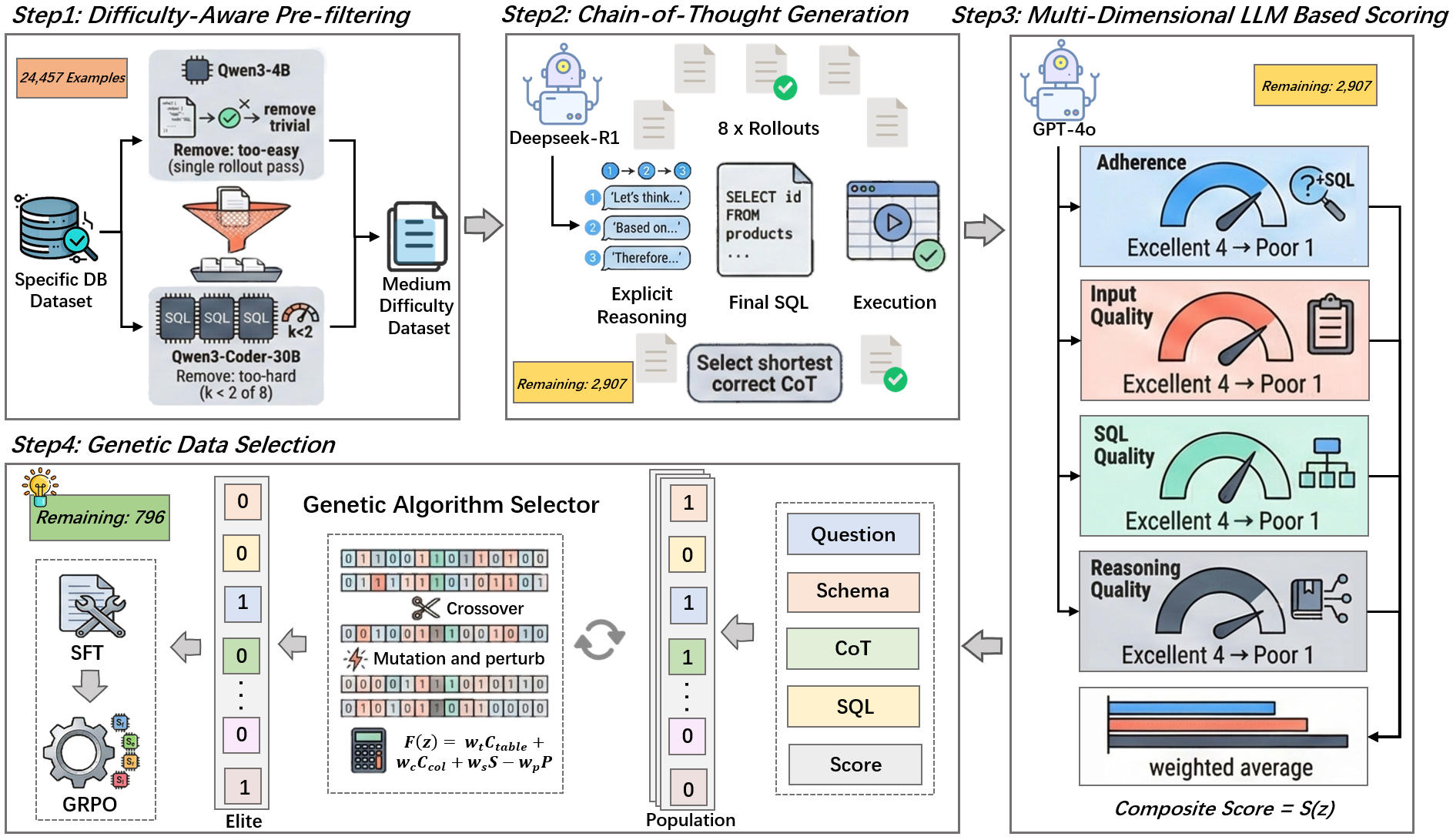}
  \caption{\label{fig1}Overall framework of LIMIT. Our framework operates through four principal phases: (1) Difficulty-Aware Pre-filtering, (2) Chain-of-Thought Generation,  (3) Multi-Dimensional LLM Based Scoring and (4) Genetic Data Selection. The data volume in the figure represents the amount of data after each stage on the BIRD dataset.}
\end{figure*}

\section{Methodology}

In this section, we introduce LIMIT, a framework for instruction tuning in Text-to-SQL that emphasizes high quality, small-scale training data and efficient model learning.
As illustrated in Figure \ref{fig1}, LIMIT comprises four key stages: 1) Difficulty-Aware Pre-filtering, 2) Chain-of-Thought Generation, 3) Multi-Dimensional LLM-Based Scoring and 4) Genetic Data Selection. 

LIMIT begins by difficulty-aware pre-filtering to remove samples that are too easy or too hard, ensuring that retained examples provide strong learning signals.
Next, high-quality CoT annotations are generated and selected. To further control data quality, a unified LLM evaluates each sample along multiple dimensions, producing a composite score used for filtering and ranking in single data level.
In dataset level, a genetic algorithm searches for an optimal subset that balances schema coverage, overall quality, and dataset size under constrained budgets.
Finally, LIMIT trains models using a two-stage protocol combining SFT and RL to enforce correctness, executability, concision, and structured output.
In the following subsections, we provide a detailed description of each component of LIMIT.

\subsection{Difficulty-Aware Pre-filtering}

To construct an efficient and informative fine tuning dataset, LIMIT expands the original DB-Explore \citep{db-explore} generation pipeline by adding richer inter-table relationships and more complex query structures (including multi-table joins and aggregation operations).
Combined with the original data from DB-Explore \citep{db-explore}, it can better support database structure learning and advanced query reasoning.

On top of this candidate set, we apply a difficulty based pre-filtering stage to select high value training samples.
We consider two undesirable cases, samples that are too easy and samples that are too hard, and address them using a tiered filtering strategy with models of different capacities.

\textbf{Removing easy samples.} 
For each sample, we perform a single rollout using Qwen3-4B \citep{Qwen3} with temperature set to 0.
If the generated SQL is correct, the sample is regarded as trivial and removed.

\textbf{Removing hard samples.} 
For the remaining samples, we conduct 8 rollouts using Qwen3-Coder-30B-A3B-Instruct \citep{Qwen3} with temperature set to 0.8.
Let $k$ be the number of correct generations. If $k < 2$, the sample is removed as overly difficult, indicating unstable or uninformative learning signals even for strong models.
This threshold preserves learnable challenges while filtering out noise dominated cases.

\subsection{Chain-of-Thought Generation}

After difficulty based pre-filtering, LIMIT further augments the retained samples with high quality CoT annotations to improve instruction fine tuning effectiveness.
We use the reasoning model DeepSeek-R1 \citep{deepseek-r1} to generate CoT for each sample. Given a natural language question and its database schema, DeepSeek-R1 \citep{deepseek-r1} is prompted under a fixed template to produce reasoning traces and the final SQL.
For each sample, we perform 8 independent rollouts to obtain diverse candidate reasoning paths.

To ensure correctness and instructional value, LIMIT applies a consistency based selection strategy.
A generated CoT is considered valid only if the execution result of its SQL is correct.
When multiple correct rollouts exist, we select the CoT with the shortest token length as the final annotation.
This preference for concise and complete reasoning reduces redundancy, highlights core reasoning patterns, and controls the overall token budget, thereby improving fine-tuning efficiency.

\subsection{Multi-Dimensional LLM Based Scoring}

After chain-of-thought generation, LIMIT introduces a model based data evaluation module to enforce fine grained and controllable quality selection. 

Building on the findings of OmniSQL \citep{omnisql}, which show that strong LLMs can approximate human judgment in Text-to-SQL data evaluation, LIMIT extends these principles by incorporating an explicit assessment of CoT quality into its scoring framework.
It employs a multi-dimensional evaluation strategy for each sample, which comprehensively considers the tight coupling between the question, SQL, and the reasoning.
We use GPT-4o \citep{Gpt-4o} as a unified evaluator to assess each sample along four complementary dimensions:

\begin{itemize}
  \item \textbf{Adherence} evaluates whether the generated SQL accurately reflects the semantics and intent of the natural language question.

  \item \textbf{Input quality} assesses the clarity, formality, and information sufficiency of the natural language question. 

  \item \textbf{SQL quality} measures the soundness of the SQL query such as structural clarity, readability and execution efficiency.

  \item \textbf{Reasoning quality} evaluates whether the generated CoT provides a clear, complete, and pedagogically useful reasoning process.
\end{itemize}

Each dimension is further subdivided into several smaller dimensions.
The evaluation prompts are provided in Appendix \ref{appendix-b-1}.

For each dimension, GPT-4o assigns a categorical rating from \textit{Excellent}, \textit{Good}, \textit{Average}, to \textit{Poor}, mapped to scores 4, 3, 2, and 1, and a weighted average over the four dimensions is then computed to produce a single composite quality score for each sample.
This score serves as a fitness signal for subsequent dataset optimization, enabling LIMIT to retain highly instructive samples under constrained data budgets.

\subsection{Genetic Data Selection} 

From a dataset level perspective, LIMIT introduces a genetic algorithm based optimization module to perform secondary filtering and combinatorial optimization over candidate samples.
Unlike previous stages that focus on individual sample quality, this stage aims to automatically search for a training subset that achieves an optimal balance among database coverage, overall quality, and dataset size under a fixed data budget.

LIMIT formulates dataset selection as a combinatorial optimization problem. Let the post filtered data pool be $D = \{d_1, \ldots, d_N\}$. Each candidate subset is represented by a binary vector $\mathbf{z} \in \{0,1\}^N$, where $z_i = 1$ indicates that sample $d_i$ is selected.
The genetic algorithm iteratively evolves a population of such vectors by optimizing a global fitness function, defined in Eq. \eqref{fitness} as a weighted combination of coverage, quality, and size control objectives. 


\begin{equation}
\label{fitness}
\begin{split}
\mathcal{F}(\mathbf{z}) &=
w_{\text{table}} C_{\text{table}}(\mathbf{z}) +
w_{\text{column}} C_{\text{column}}(\mathbf{z}) \\
&\quad + w_{\text{score}} S(\mathbf{z})
- w_{\text{penalty}} P(\mathbf{z})
\end{split}
\end{equation}

\textbf{Database coverage.}
Coverage measures how well the selected subset represents the database schema.
Table coverage $C_{\text{table}}$ is defined as the proportion of distinct tables appearing in the selected samples relative to all tables, and column coverage $C_{\text{column}}$ is defined analogously for columns.
This term encourages broad schema coverage and prevents over concentration on a small subset of tables or columns.

\textbf{Overall quality score.}
$S(\mathbf{z})$ denotes the aggregated quality score of the selected samples, obtained from the multi dimensional GPT-4o based evaluation in the previous stage.
It measures the average quality of the selected subset and serves as the main optimization signal. 

\textbf{Data size penalty.}
To enforce the Less-Is-More principle and avoid trivial improvements through increasing data volume, LIMIT introduces an explicit size penalty term $P(\mathbf{z})$.
When the number of selected samples exceeds a predefined target size, the excess is penalized proportionally, encouraging the algorithm to favor compact yet high quality subsets.

\textbf{Genetic operators.}
Crossover exchanges corresponding bits between two parent vectors to combine structural information from different subsets.
Mutation flips individual bits to introduce randomness and avoid premature convergence.
In a large binary chromosome, crossover and bit-flip mutation treat 0 and 1 symmetrically, making it difficult for the population to escape local optima in terms of dataset size.
To address this issue, LIMIT introduces cardinality mutation, which more effectively explores solutions with different dataset sizes by randomly selecting a subset of individuals and directly increasing or decreasing the number of active bits through perturbing a fraction of selected samples. 
This mechanism allows the search process to explore solutions with varying sizes and works jointly with the size penalty to identify subsets that balance quality, coverage, and scale.

The overall process and parameter settings for the optimization algorithm are detailed in the Appendix \ref{appendix-b-3}.

\subsection{Model Training} 

LIMIT adopts a two stage training paradigm combining SFT and RL to fully exploit high quality, small scale instruction data.
Prior work has shown that this paradigm improves reasoning and generation stability on Text to SQL tasks, for example SQL-R1 \citep{sql-r1}.

\textbf{Supervised fine tuning}
The SFT stage learns a mapping from $\langle Q_i, D_i\rangle$ to an executable SQL $R_i$. We fine tune the base model by minimizing the negative log likelihood
\begin{equation}
\label{generation}
Loss = - \frac{1}{|T|} \sum_{i=1}^{|T|}p(R_i | Q_i, D_i, M_i)
\end{equation}
where $Q_i$ is the natural language question, $D_i$ is the database schema, and $M_i$ denotes associated metadata.

\textbf{Reinforcement learning}
The RL stage refines generation with a composite reward that balances output format, executability, correctness, and brevity.
Referencing the successful experience of SQL-R1 \citep{sql-r1}, the overall reward is
\begin{equation}
\mathcal{R} = S_{f} + S_{e} + S_{r} + S_{l},
\end{equation}
where $S_{f}$, $S_{e}$, $S_{r}$ and $S_{l}$ denote format, execution, result, and length rewards respectively.

Training and inference prompt template is provided in Appendix \ref{appendix-b-2}.

\section{Experiments}

\begin{table*}[ht]
  \centering
  \small
  \begin{tabular}{lcc}
    \hline
    Methods                       & BIRD-Dev-EX  & Spider-Dev-EX \\
    \hline
    \multicolumn{2}{l}{\textit{Prompting with Closed-Source LLMs}} \\
    MCS-SQL + GPT-4 \citep{mac-sql}     & 63.4  & 89.5 \\
    XIYAN \citep{XiYan-SQL}      & 73.3   & 69.7 \\
    CHASE-SQL + Gemini-1.5 \citep{chase-sql}      & 73.1  & 87.6 \\
    MAC-SQL + GPT-4 \citep{mac-sql}     & 59.4  & 86.8 \\
    Agentar-Scale-SQL \citep{AgentarScaleSQLAT}     & 74.9  &- \\
    \hline
    \multicolumn{2}{l}{\textit{Fine-Tuning with Open-Source LLMs 14B-15B}} \\
    CODES-15B \citep{codes}      & 58.5  &84.9 \\
    ROUTE + Qwen2.5-14B \citep{ROUTE}      & 60.9   &87.3 \\
    OmniSQL-14B \citep{omnisql}      & 64.2  &81.4 \\ 
    SQL-R1 + Qwen2.5-Coder-14B(self-consistency@8) \citep{sql-r1}     & 67.1  &86.7 \\
    Reasoning-SQL + Qwen2.5-Coder-14B\citep{reasoning-sql}     & 65.3  &81.4 \\
    \hline
    \multicolumn{2}{l}{\textit{Fine-Tuning with Open-Source LLMs 7B-8B}} \\
    CODES-7B \citep{codes}      & 57.2  & 85.4 \\
    ROUTE + Qwen2.5-7B \citep{ROUTE}      & 55.9  & 83.6 \\
    OmniSQL-7B \citep{omnisql}      & 63.9   & 81.6 \\ 
    SQL-o1 + Qwen2.5-7B \citep{sql-o1}      & 66.7  & 84.7 \\
    SQL-o1 + Llama3-8B \citep{sql-o1}      & 63.4   &87.4 \\
    SQL-R1 + Qwen2.5-Coder-7B(self-consistency@8) \citep{sql-r1}     & 66.6  & 87.6 \\
    Reasoning-SQL + Qwen2.5-Coder-7B\citep{reasoning-sql}     & 64.0   & - \\
    SLM-SQL(self-consistency@16)\citep{slm-sql}     & \underline{67.1}  &76.7 \\
    EvolSQL + Qwen2.5-Coder-7B\citep{evol-sql}              &65.1             &86.1 \\
    RingSQL + Qwen2.5-Coder-7B\citep{ring-sql}             &61.7            &85.5 \\
    SQL-Trail + Qwen2.5-Coder-7B(self-consistency@8) \citep{sql-trail}            &64.2       &86.8 \\       
    DB-Explore \citep{db-explore}      & 65.2  & 87.2 \\
    DB-Explore + self-consistency@8 \citep{db-explore}     & 67.0  & 87.8 \\

    \hline
    \textbf{Ours: LIMIT + Qwen3-8B}     & 66.9  &\underline{88.4} \\
    \textbf{Ours: LIMIT + Qwen3-8B + self-consistency@8}     & \textbf{69.1}   &\textbf{88.9} \\
    \hline
  \end{tabular}
    \caption{\label{section-4-main}
    Experimental results on BIRD benchmarks.
    Within the Fine-Tuning with Open-Source LLMs 7B-8B group, top-performing entries are \textbf{bold} with runner-up scores \underline{underlined}.
  }
\end{table*}

\paragraph{Benchmarks}  


We evaluate LIMIT on the BIRD benchmark \citep{bird} and the Spider dataset \citep{spider}.
BIRD contains 95 large scale relational databases from real world business systems across 37 domains, with highly complex schemas and realistic Text to SQL challenges.
Spider is a large scale cross domain Text to SQL benchmark with $10{,}181$ questions and $5{,}693$ unique complex SQL queries over 200 databases spanning 138 domains.
We use execution accuracy (EX) \citep{spider} as the evaluation metric.
EX compares query results by executing generated SQL against the reference outputs, avoiding sensitivity to syntactic variations and providing a reliable measure of practical correctness in real database environments.

\paragraph{Implementation Details}  

LIMIT uses Qwen3-8B \citep{Qwen3} as the base model.

The initial DB-Explore pool for BIRD dev grew from 16,457 to 24,457 samples; after pre-filtering 2,907 candidates remained and the evaluation and optimization modules yielded 796 high quality training samples.
The Spider pool grew from 21,237 to 29,237, was reduced by pre-filtering to 3,476, and produced 863 final samples.
For a detailed training cost analysis, please refer to Appendix \ref{appendix-b-5}.

Training is implemented with LLaMA-Factory \citep{Llamafactory} and VERL \citep{verl}. In the SFT stage we use the AdamW optimizer for 15 epochs with an initial learning rate of $1\times10^{-5}$ and a cosine decay schedule.
The context window is set to 4096 tokens.
In the RL stage we apply a GRPO based policy optimizer with learning rate $3\times10^{-7}$ and 4 rollouts per sample.
For inference we use self-consistency decoding. Under $\tau=0.8$ we sample $8$ candidate SQLs per input and select the final output via execution consistency. For single shot generation we set temperature to 0 to ensure deterministic output.
Genetic algorithm parameters and sensitivity analysis are shown in Appendix \ref{appendix-b-4}.

\paragraph{Baselines}  


We benchmark LIMIT against a diverse set of recent Text to SQL methods.
MCS-SQL \citep{mcs-sql} uses multiple prompts and result selection to reduce single shot instability,
XIYAN \citep{XiYan-SQL} leverages structured reasoning and tool enhanced prompting,
CHASE-SQL \citep{chase-sql} applies multi round reasoning with candidate tracking.
Agentar-Scale-SQL \citep{AgentarScaleSQLAT} employs orchestrated test time scaling with RL enhanced internal reasoning and parallel expansion.
MAC-SQL \citep{mac-sql} uses multi agent collaboration for generation validation and correction.
CODES \citep{codes} integrates large scale code pretraining and plugins,
ROUTE \citep{ROUTE} applies multi task learning,
OmniSQL \citep{omnisql} uses large scale instruct fine tuning,
SQL-o1 \citep{sql-o1} and SQL-R1 \citep{sql-r1} leverage execution-guided reasoning tuning; Reasoning-SQL \citep{reasoning-sql}, SLM-SQL \citep{slm-sql}, and SQL-TRAIL \citep{sql-trail} apply RL-based reasoning strategies; EVOLSQL \citep{evol-sql} and RINGSQL \citep{ring-sql} focus on scalable synthetic Text-to-SQL data generation.

\section{Results and Analysis}

\subsection{Main Results}

\begin{figure}[h]
  \includegraphics[width=\columnwidth]{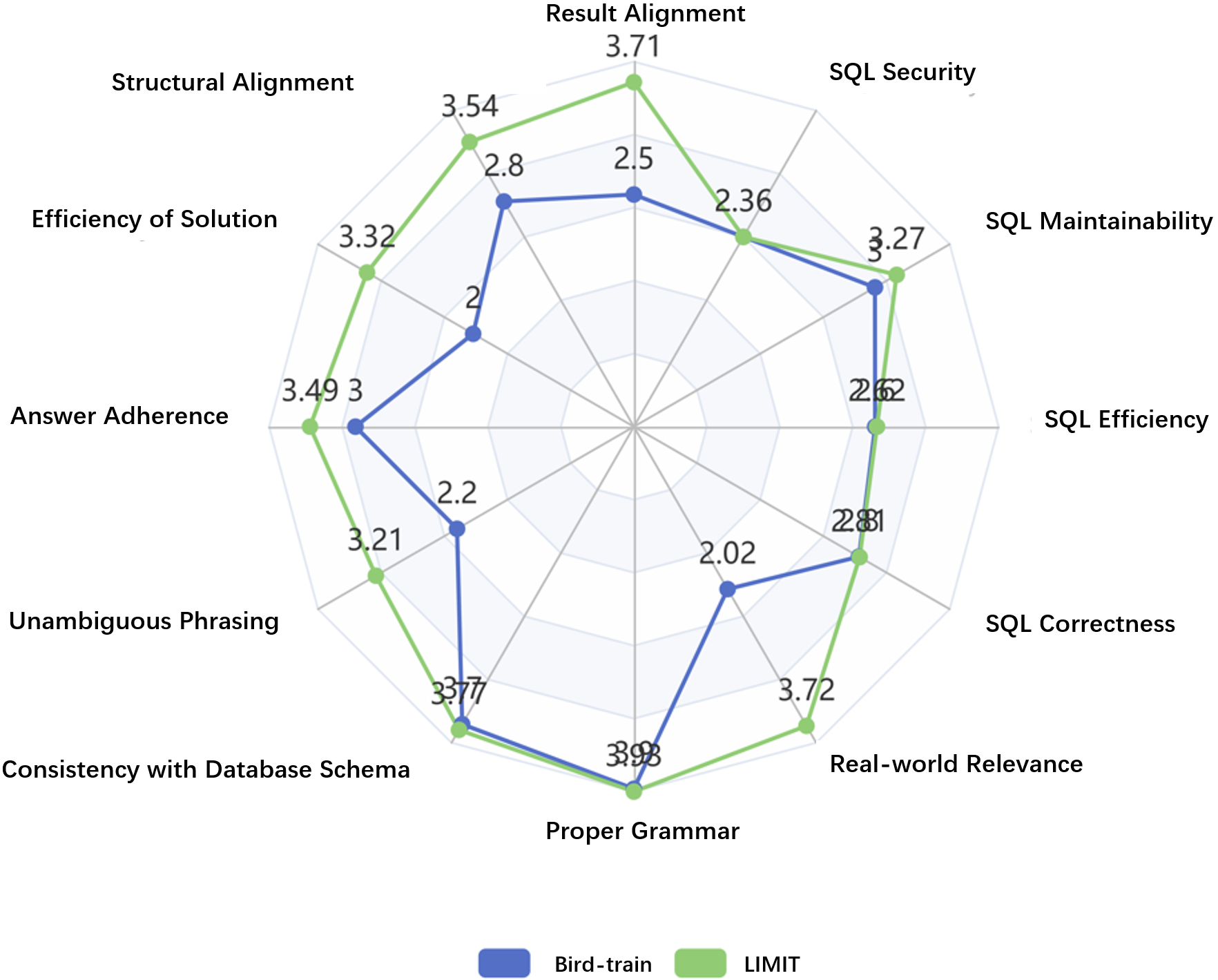}
  \caption{\label{fig2}Comparison of dataset scores between LIMIT and BIRD train.}
\end{figure}

\begin{table}[ht]
  \centering
  \small
  \begin{tabular}{lc}
    \hline
    Methods                       & BIRD-Dev-EX \\
    \hline
    Qwen3-8B \citep{Qwen3}     & 57.4\\
    \hline
    SFT     & 65.4 \\
    SFT + RL     & 66.9\\
    SFT + RL + self-consistency@8     & 69.1\\

    \hline
  \end{tabular}
    \caption{\label{section-4-rl}
    Performance of LIMIT at different training stages on the BIRD dataset.
}
\end{table}

\begin{table}[ht]
  \centering
  \small
  \begin{tabular}{lc}
    \hline
    Methods                    & {Data volume}    \\
    \hline
    Bird-Train & 9428 \\
    Codes \citep{codes} & 21.5GB \\
    OmniSQL \citep{omnisql} & 2500000 \\
    Route \citep{ROUTE} & 30000 \\
    DB-Explore \citep{db-explore} & 16457 \\
    LIMIT & 796\\
    \hline
  \end{tabular}
  \caption{\label{section4-volume}
    LIMIT data volume analysis in BIRD.
}
\end{table}

\begin{table}[htb]
  \centering
  \small
  \begin{tabular}{lcc}
    \hline
    Cover Rate                       & {Table}  & {column}   \\
    \hline
    Bird-DEV & 100\% & 71.7\% \\
    DB-Explore \citep{db-explore} & 100\% & 100\% \\
    LIMIT & 100\% & 86.8\% \\
    \hline
  \end{tabular}
  \caption{\label{section4-db}
    DB coverage analysis.
}
\end{table}

\paragraph{Performance on Main Benchmarks}

\begin{figure}[h]
  \includegraphics[width=\columnwidth]{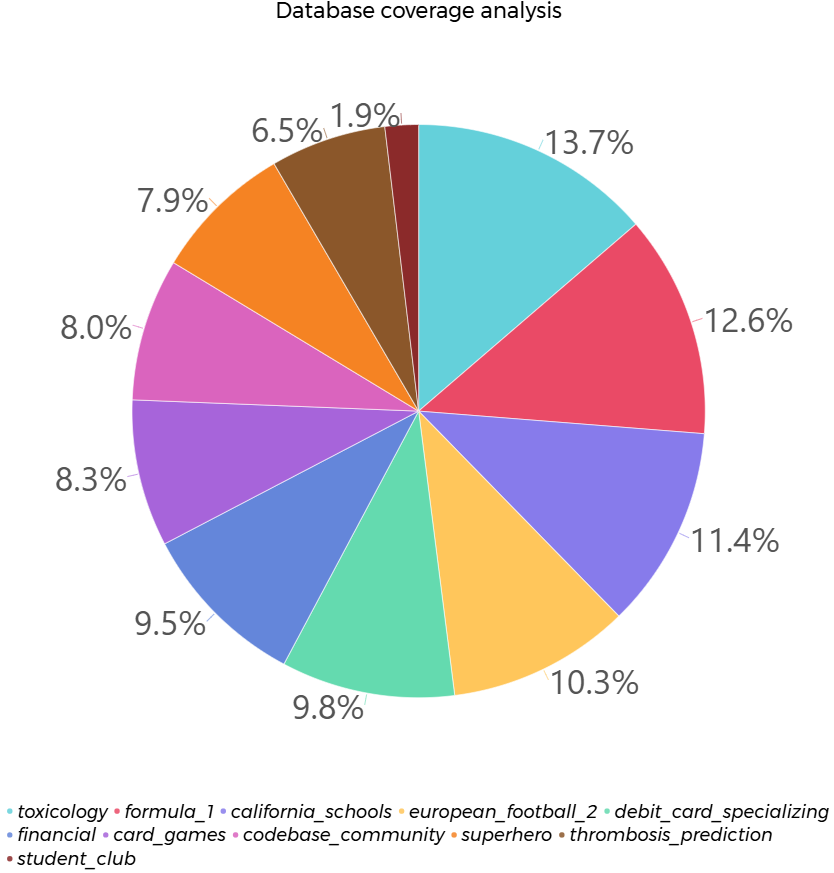}
\caption{\label{fig3}LIMIT DB Coverage Analysis.}
\end{figure}

Table \ref{section-4-main} summarizes the performance of LIMIT on BIRD-Dev-EX and Spider-Dev-EX, comparing it with representative prompting based approaches and fine tuned open source models.
Overall, LIMIT achieves competitive and often superior performance with a very small amount of data. Using only an 8B open source model, it surpasses multiple methods based on larger 14B models and remains competitive with strong closed source baselines, validating the Less-Is-More principle for Text to SQL.
Among fine tuning methods with 7B to 8B models, SQL-o1 \citep{sql-o1}, SQL-R1 \citep{sql-r1}, and SLM-SQL \citep{slm-sql} represent state of the art reasoning enhanced systems with peak BIRD accuracy around 66\% to 67\%.
DB-Explore \citep{db-explore}, which forms the basis of LIMIT, reaches 67.0\% with self consistency, demonstrating the effectiveness of database exploration and instruction synthesis.
Building on this foundation, LIMIT applies strict data selection and combines SFT with RL, achieving 66.9\% under single pass inference with fewer but more focused samples, indicating that large scale instruction data is not necessary for strong performance.
With self consistency@8 \citep{Self-Consistency}, LIMIT further improves to 69.1\%, outperforming DB-Explore with self consistency and strong baselines such as SQL-o1 and SLM-SQL.
On Spider, LIMIT attains 88.4\% EX and reaches 88.9\% with self consistency, outperforming most open source fine tuned models and remaining competitive with closed source methods, which suggests that LIMIT generalizes well across datasets and better supports stable reasoning under multi path sampling.

\paragraph{Analysis of Training Stage}

\begin{table}[htb]
  \centering
  \small
  \begin{tabular}{lcc}
    \hline
    Methods                       & \multicolumn{2}{c}{BIRD-Dev-EX}    \\
    ~                               & SFT          & RL     \\
    \hline
    \textbf{Ours: LIMIT + Qwen3-8B }    & \textbf{65.4}              & \textbf{66.9}\\

    - w/o DB Cover Rate & 58.4$_{\downarrow7.0}$ & 61.1 $_{\downarrow5.8}$ \\
    - w/o Adherence& 60.6$_{\downarrow4.8}$ & 61.4$_{\downarrow5.5}$     \\
    - w/o Input Quality & 63.8$_{\downarrow1.6}$ & 65.0$_{\downarrow1.9}$ \\
    - w/o SQL Quality  & 63.5$_{\downarrow1.9}$ & 64.8$_{\downarrow2.1}$ \\
    - w/o Reasoning Quality  & 64.1$_{\downarrow1.3}$ & 64.1$_{\downarrow2.8}$ \\
    \hline
  \end{tabular}
  \caption{\label{section4-ablation}
    Ablation study on LIMIT, complete method of LIMIT is \textbf{bold}.
}
\end{table}

Table \ref{section-4-rl} reports how model performance on BIRD-Dev-EX evolves across training stages, allowing a systematic analysis of data selection, SFT and RL.
Qwen3-8B \citep{Qwen3} achieves 57.4\% accuracy on BIRD \citep{bird}, which indicates reasonable general language understanding but limited ability to produce semantically correct and executable SQL on real world databases with complex schemas and implicit constraints. 
Fine-tuning with 796 high quality instruction samples selected by LIMIT improves accuracy to 65.4\%, an increase of 8.3 percentage points.
Introducing the RL stage further raises performance to 66.9\%, highlighting the high quality of the generated CoT, as well as the high information density and instructional value of the curated data. 
Finally, applying self consistency@8 increases accuracy to 69.1\%, which demonstrates that the selected training examples offer diverse and coherent reasoning paths that allow multi sample voting to converge more reliably to correct SQL.
Together these results emphasize the importance of domain aligned, high quality samples for efficient Text-to-SQL learning and the complementary roles of SFT, RL and multi sample aggregation.

\paragraph{Analysis of Data Quality}

To analyze the advantages of LIMIT in data quality and data efficiency, we compare it with representative Text-to-SQL methods from the perspectives of data scale, multi dimensional quality scores, and database coverage.
Table \ref{section4-volume} shows that most prior approaches rely on large scale training data.
In contrast, LIMIT selects only 796 high quality samples from the DB-Explore pool, reducing data size by about 95\% relative to DB-Explore and by several orders of magnitude compared to OmniSQL, while still achieving competitive performance.
This result supports the Less-is-More principle, where performance gains depend more on semantic alignment, structural coverage, and reasoning signals than on raw data volume.
Figure \ref{fig2} further shows that LIMIT consistently outperforms the original BIRD \citep{bird} training set in overall quality scores, especially in instruction clarity, question SQL alignment, and reasoning consistency, while remaining comparable on engineering oriented aspects such as SQL security, maintainability, efficiency, and correctness.
Table \ref{section4-db} reports that LIMIT achieves full table coverage and 86.8\% column coverage with only 796 samples, striking a balance between coverage and compactness, whereas BIRD has lower column coverage and DB-Explore achieves full coverage at much higher cost.
Figure \ref{fig3} shows that LIMIT samples are evenly distributed across 11 databases, avoiding concentration on a few schemas and favoring databases with higher structural complexity and reasoning value.

\subsection{Ablation Study}

Table \ref{section4-ablation} reports module level ablation results of LIMIT on BIRD-Dev-EX at the SFT and RL stages.
The full framework achieves 65.4\% after SFT and 66.9\% after RL, serving as the performance upper bound.
Removing the DB cover rate constraint causes the largest degradation, with accuracy dropping to 58.4\% at SFT and decreasing by 5.8 points at RL, showing that insufficient table and column coverage leads to overfitting to local schemas and weak cross table generalization on complex benchmarks such as BIRD \citep{bird}.
Removing the adherence constraint also yields substantial drops of 4.8\% and 5.3\%, indicating that misalignment between questions and SQL structures harms the learning of correct mappings.
Input quality, SQL quality, and reasoning quality have milder but consistent effects.
Excluding input quality reduces performance by about 1.6\% and 1.9\%, while removing SQL quality leads to larger drops, especially at RL, reflecting the importance of executable and efficient queries under execution based rewards.
Notably, removing reasoning quality causes the largest RL stage drop of 2.8\%, exceeding the SFT drop, which shows that high quality reasoning traces are particularly critical for RL.
Overall, these results demonstrate that LIMIT relies on the synergy of schema coverage and multiple quality constraints, with DB coverage and adherence as core factors and reasoning quality amplifying gains during RL.

\section{Conclusion}

We introduce LIMIT, a data centric pipeline for Text-to-SQL instruction tuning that pioneeringly applies the Less-Is-More principle to this domain.
LIMIT emphasizes compact, high quality datasets over large instruction corpora by integrating difficulty-aware pre-filtering, chain-of-thought generation, multi-dimensional LLM based scoring, and genetic data selection, followed by SFT and RL.
On the BIRD benchmark, LIMIT achieves competitive performance with only 796 curated samples.
Overall, LIMIT offers a practical and extensible path toward data efficient Text-to-SQL learning across larger models and broader domains. 

\section*{Limitations}

Although LIMIT demonstrates strong data efficiency and highly competitive performance, several limitations remain.
First, while the framework significantly reduces training-time computation, it relies on multiple large language models for CoT generation and multi-dimensional scoring, introducing additional API costs that may hinder reproducibility.
Second, although LIMIT achieves state-of-the-art results among approaches based on 7B-8B open-source models, it still lags behind some methods built on substantially larger models.
Finally, while LIMIT performs well on an 8B backbone, its effectiveness on much larger model architectures has not been systematically studied, leaving its scalability and robustness across different model sizes as open questions.

\section*{Ethics Statements}

We use LLMs to synthesize training data and to perform data quality evaluation, as well as to generate chain-of-thought annotations. All processes are conducted exclusively on publicly available Text-to-SQL benchmarks and database schemas. No private, proprietary, or personal data are used, and no human subjects are involved.

\bibliography{main} 

\clearpage

\appendix

\section{LLM based scoring rules} \label{appendix-b-1}
\begin{figure*}[htb]
    \centering
    \includegraphics[width=1\linewidth]{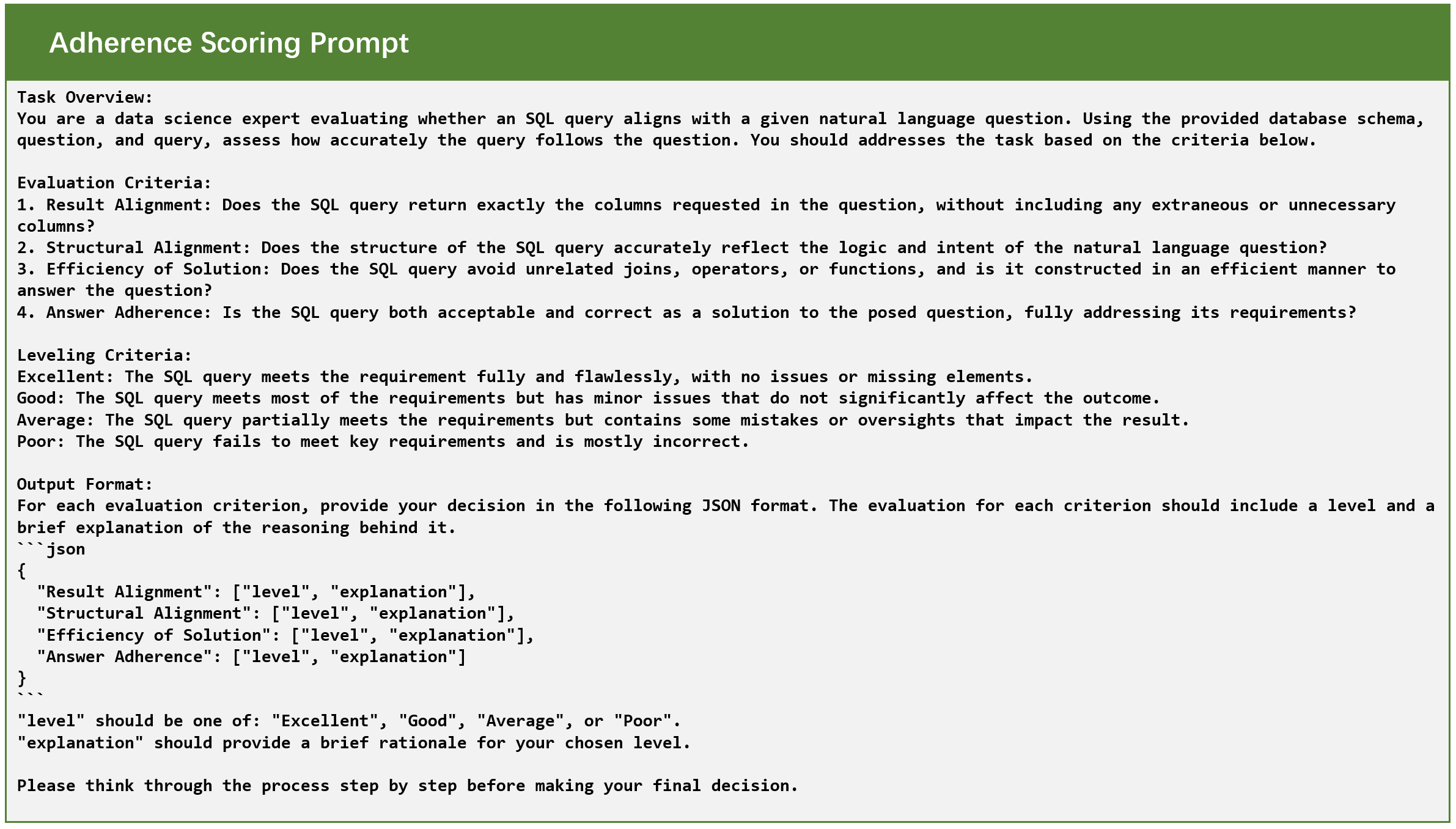}
    \caption{Adherence scoring prompt.}
    \label{appendix-b1}
\end{figure*}

\begin{figure*}[htb]
    \centering
    \includegraphics[width=1\linewidth]{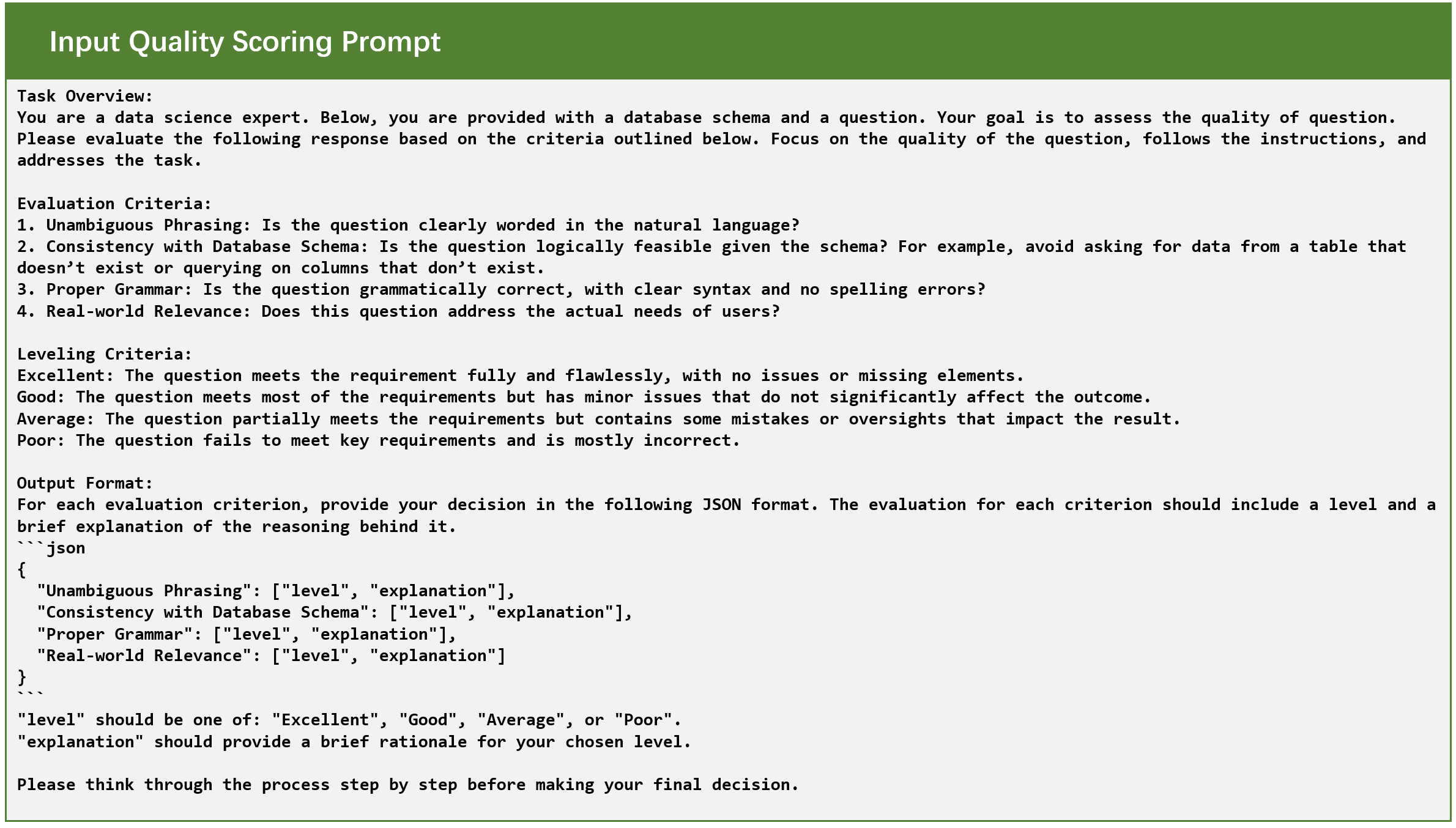}
    \caption{Input quality scoring prompt.}
    \label{appendix-b2}
\end{figure*}

\begin{figure*}[htb]
    \centering
    \includegraphics[width=1\linewidth]{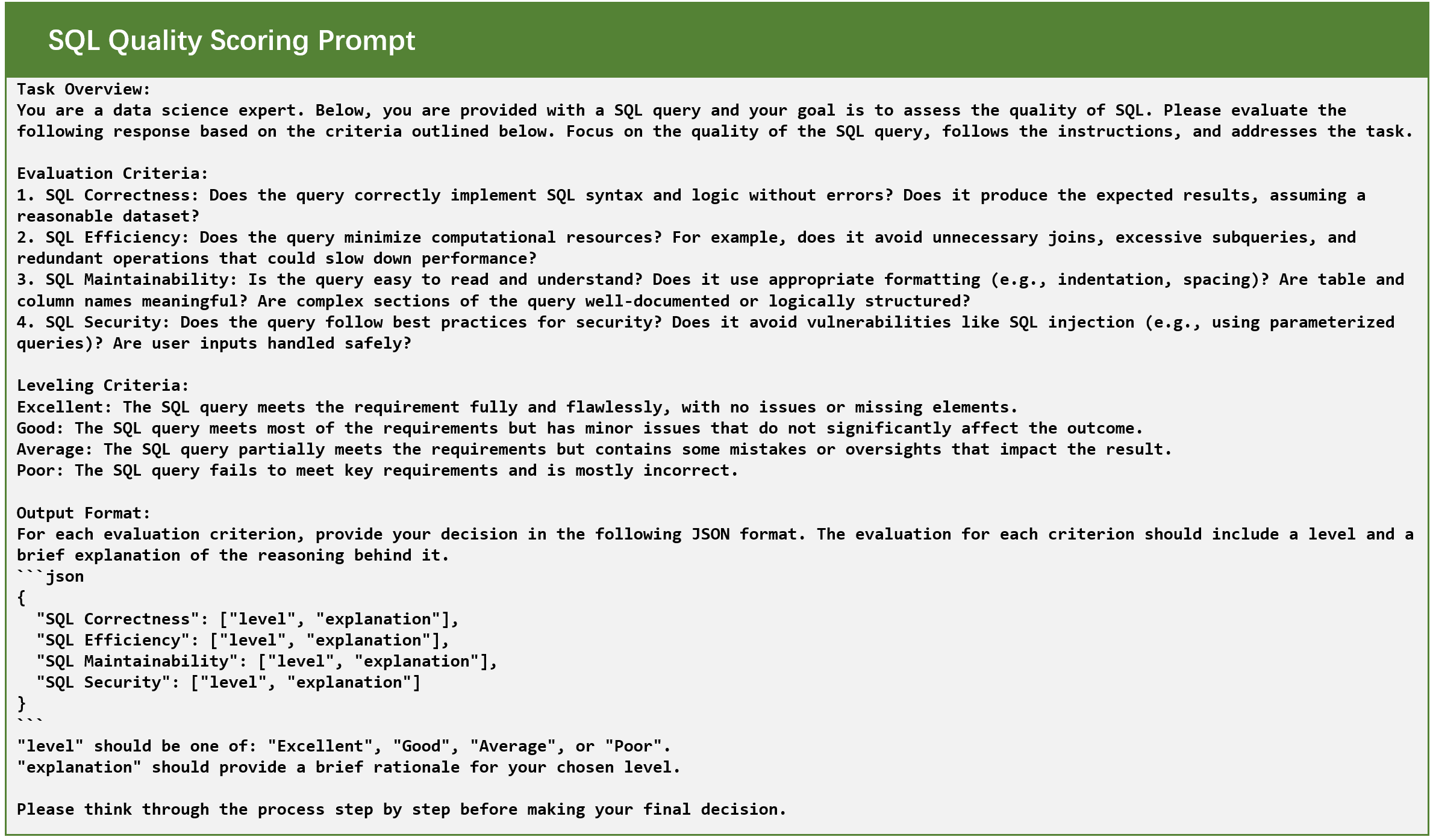}
    \caption{SQL quality scoring prompt.}
    \label{appendix-b3}
\end{figure*}

\begin{figure*}[htb]
    \centering
    \includegraphics[width=1\linewidth]{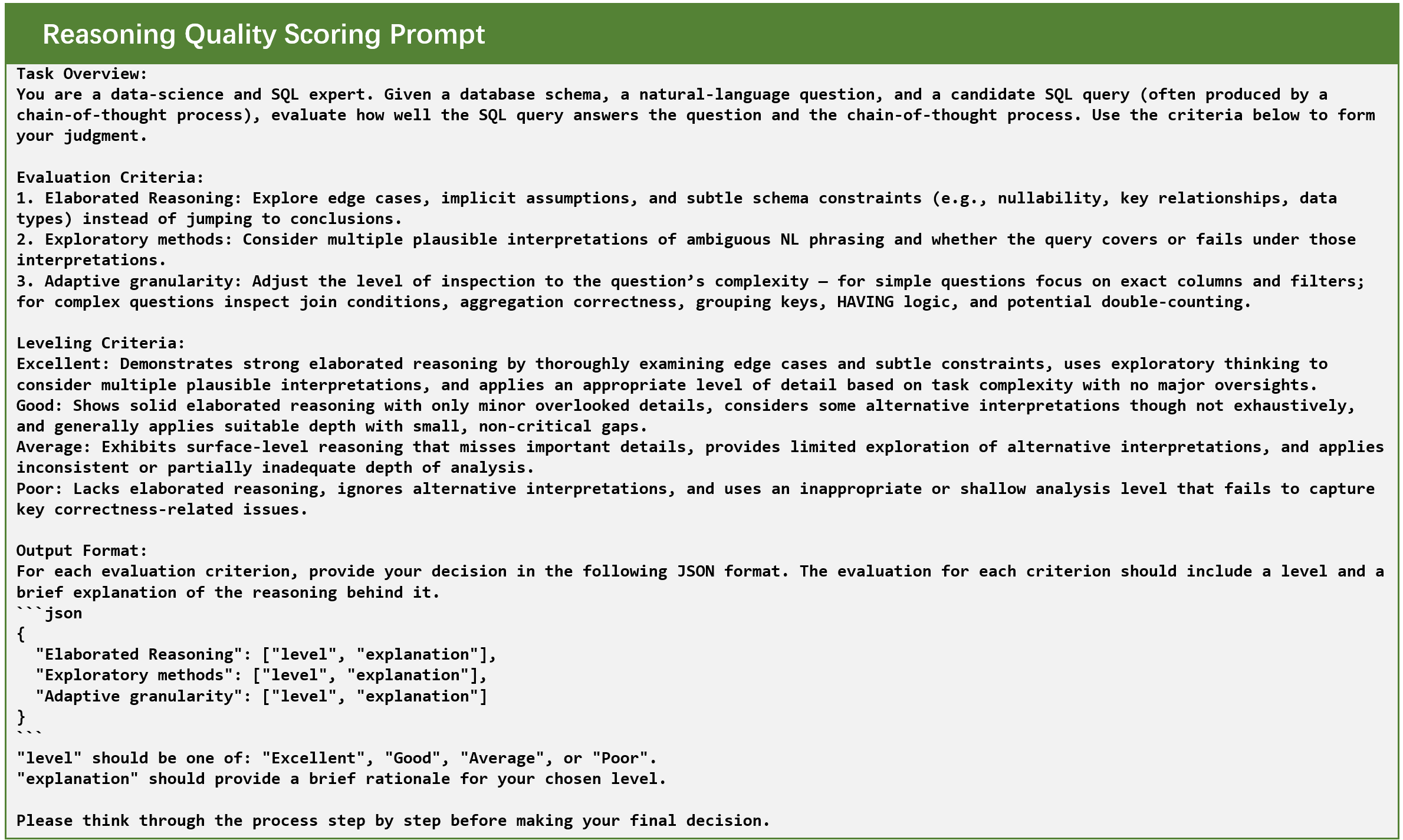}
    \caption{Reasoning quality scoring prompt.}
    \label{appendix-b4}
\end{figure*}

The scoring rules for LIMIT are as shown in Figure \ref{appendix-b1} - \ref{appendix-b4}

\section{LIMIT inference prompt} \label{appendix-b-2}
\begin{figure*}[htb]
    \centering
    \includegraphics[width=1\linewidth]{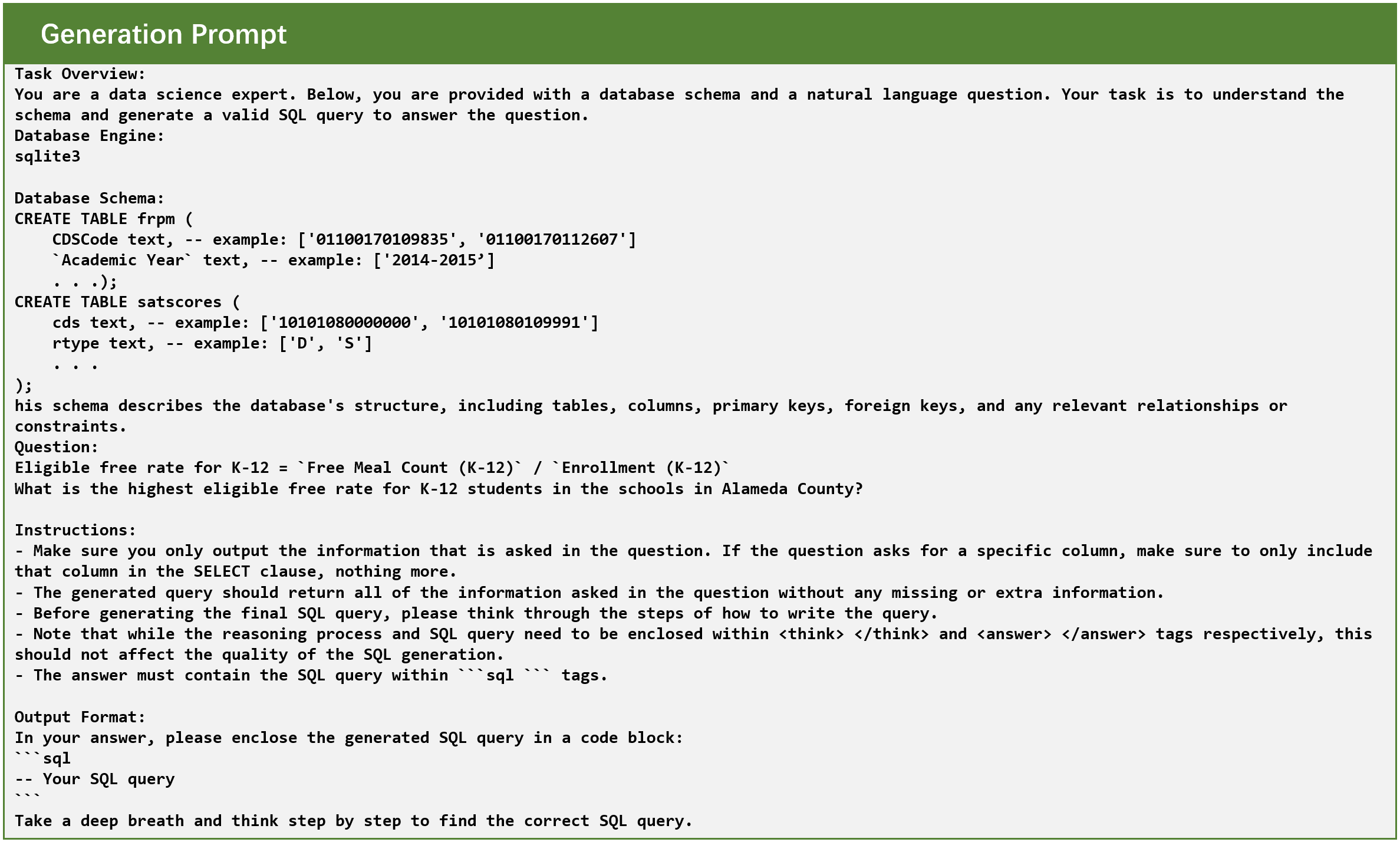}
    \caption{LIMIT inference prompt.}
    \label{appendix-b5}
\end{figure*}

The prompt used in LIMIT inference phase are as shown in Figure \ref{appendix-b5}

\section{Genetic Data Selection Algorithm} \label{appendix-b-3}
The overall process of the Genetic Data Selection Algorithm is shown in Algorithm \ref{optimization}.

\paragraph{Fitness Function.}
Given a selection vector $\mathbf{z}\in\{0,1\}^N$, let
\begin{equation}
k=\sum_{i=1}^N z_i
\end{equation}
denote the number of selected samples. The fitness components are defined as
\begin{align}
C_{\text{table}} &= \frac{|\mathcal{T}_{\text{sel}}|}{T_{\text{total}}}, \\
C_{\text{column}} &= \frac{|\mathcal{C}_{\text{sel}}|}{C_{\text{total}}}, \\
S &= \frac{1}{4k}\sum_{i:z_i=1}\text{score}_i, \\
P &= \frac{\max(0,k-T)}{T}.
\end{align}

The overall fitness is computed as

\begin{equation}
\label{fit}
\begin{split}
\mathcal{F}(\mathbf{z}) &=
w_{\text{table}} C_{\text{table}}(\mathbf{z}) +
w_{\text{column}} C_{\text{column}}(\mathbf{z}) \\
&\quad + w_{\text{score}} S(\mathbf{z})
- w_{\text{penalty}} P(\mathbf{z})
\end{split}
\end{equation}

Specifically, we assign $w_{\text{table}}=0.2$ and $w_{\text{column}}=0.3$ to encourage broad table and column coverage, $w_{\text{score}}=0.5$ to prioritize high quality samples, and $w_{\text{penalty}}=0.5$ to impose a strong constraint on the selected subset size.

\section{Genetic Data Selection sensitivity analysis} \label{appendix-b-4}

\begin{table*}[h]
\centering
\small
\begin{tabular}{lccccc}
\toprule
Weight & Value & Subset Size & Table Cov. (\%) & Column Cov. (\%) & EX (\%) \\
\midrule

\multirow{4}{*}{$w_{\text{table}}$}
 & 0.2 (Default) & 796 & 100.0 & 86.8 & \textbf{66.9} \\
 & 0.1 & 761 & 99.2 & 83.5 & 65.1 \\
 & 0.3 & 828 & 100.0 & 88.5 & 66.1 \\
 & 0.4 & 912 & 100.0 & 91.4 & 65.6 \\

\midrule

\multirow{4}{*}{$w_{\text{column}}$}
 & 0.3 (Default) & 796 & 100.0 & 86.8 & \textbf{66.9} \\
 & 0.1 & 785 & 100.0 & 78.2 & 64.3 \\
 & 0.2 & 790 & 100.0 & 82.5 & 65.4 \\
 & 0.4 & 847 & 100.0 & 91.0 & 65.2 \\

\midrule

\multirow{4}{*}{$w_{\text{score}}$}
 & 0.5 (Default) & 796 & 100.0 & 86.8 & \textbf{66.9} \\
 & 0.3 & 819 & 100.0 & 87.5 & 64.7 \\
 & 0.4 & 808 & 100.0 & 87.0 & 66.0 \\
 & 0.6 & 743 & 99.2 & 84.0 & 65.5 \\

\midrule

\multirow{4}{*}{$w_{\text{penalty}}$}
 & 0.5 (Default) & 796 & 100.0 & 86.8 & \textbf{66.9} \\
 & 0.3 & 845 & 100.0 & 88.2 & 65.7 \\
 & 0.4 & 822 & 100.0 & 87.4 & 66.2 \\
 & 0.6 & 675 & 98.7 & 82.0 & 64.3 \\

\bottomrule
\end{tabular}
\caption{Sensitivity analysis of fitness weights. Each block varies one weight while fixing others at default values.}
\label{tab:sensitivity}
\end{table*}

We conducted a controlled sensitivity analysis by varying one weight at a time while keeping the other three fixed at their default values. As shown in Table \ref{tab:sensitivity}, the Default configuration consistently achieves the highest execution accuracy of 66.9, indicating that it lies in a stable near optimal region rather than being arbitrarily selected.
When adjusting the table coverage weight, decreasing it reduces both coverage and execution accuracy to 65.1, while increasing it enlarges the subset and slightly improves coverage but lowers accuracy to 66.1 and 65.6, suggesting that over emphasizing coverage introduces redundancy that harms generalization. For the column coverage weight, lowering it substantially reduces column coverage and drops accuracy to 64.3, while increasing it expands the subset but still underperforms the Default at 65.2, indicating diminishing returns from excessive diversity. For the quality score weight, both decreasing and increasing it from the Default lead to lower accuracy between 64.5 and 66.0, showing that over prioritizing either coverage or evaluator score disrupts the balance. Finally, for the penalty weight, reducing regularization inflates the subset and lowers accuracy to 65.7, whereas increasing it compresses the subset to 675 examples and further reduces accuracy to 64.3.
Overall, performance degrades smoothly when moving away from the Default configuration, demonstrating that our weight choice achieves the best tradeoff between coverage, compactness, and quality and is robust rather than arbitrarily tuned.

\section{Performance and cost analysis} \label{appendix-b-5}

\begin{table}[h]
\centering
\small
\begin{tabular}{lccc}
\toprule
Methods & EX (GRE) & EX (MV) & Samples \\
\midrule
Qwen3-8B + Full data & 67.8 & 69.6 & 24,457 \\
LIMIT & 66.9 & 69.1 & 796 \\
\bottomrule
\end{tabular}
\caption{Comparison between full data training and LIMIT.}
\label{tab:main_comparison}
\end{table}

We further analyze the computational efficiency of LIMIT by jointly considering training cost and performance (Table~\ref{tab:main_comparison}). Despite using only 796 samples, LIMIT achieves competitive execution accuracy (66.9\% vs. 67.8\% on GRE, and 69.1\% vs. 69.6\% on MV) compared to full-data training with 24K samples, indicating strong sample efficiency.

From a training perspective, under the same hardware configuration (four NVIDIA A6000 GPUs), fine-tuning Qwen3-8B with LIMIT for 10 epochs takes \textbf{5.3} hours, whereas training on the full dataset for 4 epochs requires \textbf{61.5} hours. This corresponds to an approximate \textbf{91\% reduction} in GPU training time, demonstrating a substantial improvement in computational efficiency.

Regarding data construction, LIMIT introduces additional one-time costs due to LLM-based processing. Specifically, the Chain-of-Thought generation stage consumes \textbf{11,872,188} input tokens and \textbf{5,360,508} output tokens, while the multi-dimensional scoring stage requires \textbf{4,857,597} input tokens and \textbf{4,206,429} output tokens. A significant portion of these tokens arises from incorporating complex database schema information into the prompts.

However, this upfront cost is amortized in practical scenarios. Once constructed, the curated dataset can be reused across multiple training runs, model variants, or downstream adaptations. In contrast, the reduction in training time directly translates to recurring savings in GPU computation. Therefore, LIMIT not only improves data efficiency but also leads to a more cost-effective training pipeline overall, especially in iterative or resource-constrained settings.


\begin{algorithm*}[htbp]
\caption{Genetic Data Selection Algorithm}
\label{optimization}
\begin{algorithmic}[1]
\Require Pre filtered set $D=\{d_1,\dots,d_N\}$; total tables $T_{\text{total}}$, total columns $C_{\text{total}}$; \\
Weights $w_{\text{table}},w_{\text{column}},w_{\text{score}},w_{\text{penalty}}$;  \\
Target size $T$; population size $P$, generations $G$; \\
Crossover probability $p_c$, per bit mutation probability $p_m$, cardinality mutation probability $p_{\text{cm}}$, cardinality fraction $\rho$; \\
Elite size $E$.
\Ensure Optimized subset encoding $\mathbf{z}_{\text{best}}$
\State Initialize a population of $P$ binary vectors of length $N$
\State $Q \leftarrow \emptyset$ \Comment{new population}
\For{$g=1$ to $G$}
    \State Select parents via tournament selection to form a mating pool
    \State Copy the top $E$ individuals from the current population into $Q$ \Comment{elitism}
    \For{each randomly sampled parent pair $(x,y)$ from the mating pool}
        \State $u \sim \mathcal{U}(0,1)$
        \If{$u < p_c$}
            \State Perform single point crossover on $(x,y)$ to produce offspring $(o_1,o_2)$
        \Else
            \State Copy $(o_1,o_2) \leftarrow (x,y)$
        \EndIf
        \State Add $o_1,o_2$ to $Q$
        \If{$|Q| \ge P$}
            \State \textbf{break}
        \EndIf
    \EndFor
    \For{each offspring $o$ in $Q$ excluding elites}
        \For{$i=1$ to $N$}
            \State $v \sim \mathcal{U}(0,1)$
            \If{$v < p_m$}
                \State $o_i \leftarrow 1 - o_i$
            \EndIf
        \EndFor
        \State $w \sim \mathcal{U}(0,1)$
        \If{$w < p_{\text{cm}}$}
            \State $k \leftarrow \sum_{i=1}^N o_i$
            \State $\Delta \leftarrow \lceil \rho \cdot k \rceil$
            \State Draw $b \sim \text{Bernoulli}(0.5)$
            \If{$b = 1$}
                \State Randomly set $\min(\Delta,k)$ active bits in $o$ to $0$
            \Else
                \State Randomly set $\Delta$ inactive bits in $o$ to $1$
            \EndIf
        \EndIf
    \EndFor
    \State Evaluate fitness of each individual in $Q$
    \State Set population $\leftarrow Q$
\EndFor
\State \Return the individual with highest fitness $\mathbf{z}_{\text{best}}$
\end{algorithmic}
\end{algorithm*}

\end{document}